\documentclass[10pt,twocolumn,letterpaper]{article}

\usepackage[accsupp]{axessibility}  
\usepackage{wacv}
\usepackage{times}
\usepackage{epsfig}
\usepackage{graphicx}
\usepackage{amsmath}
\usepackage{amssymb}
\usepackage{multirow}
\usepackage{times}
\usepackage{latexsym}
\usepackage[T1]{fontenc}
\usepackage[utf8]{inputenc}
\usepackage{microtype}
\usepackage{subcaption}
\usepackage{microtype}
\usepackage{enumitem}
\usepackage{rotating}
\usepackage{amsfonts}
\usepackage{amsmath}
\usepackage{soul}
\usepackage{mathtools}
\usepackage{chngcntr}
\usepackage{soul,xcolor}
\usepackage{pifont}
\usepackage{stackengine}

\newcommand{\cmark}{\textcolor{green}{\ding{51}}}%
\newcommand{\xmark}{\textcolor{red}{\ding{55}}}%

\usepackage{array,etoolbox}
\preto\tabular{\setcounter{magicrownumbers}{0}}
\newcounter{magicrownumbers}

\newcommand{\model}{\texttt{MuLOT}}

\def\wacvPaperID{1343} 

\wacvfinalcopy 

\ifwacvfinal
\fi

\ifwacvfinal
\usepackage[breaklinks=true,bookmarks=false]{hyperref}
\else
\usepackage[pagebackref=true,breaklinks=true,colorlinks,bookmarks=false]{hyperref}
\fi

\begin{document}

\title{Multimodal Learning using Optimal Transport \\ for Sarcasm and Humor Detection}

\author{Shraman Pramanick\thanks{denotes equal contribution},\; Aniket Roy\footnotemark[1],\; Vishal M. Patel \\
  Johns Hopkins University \\
  \small\texttt{\{spraman3,aroy28,vpatel36\}@jhu.edu}}

\maketitle

\ifwacvfinal
\thispagestyle{empty}
\fi

\begin{abstract}
Multimodal learning is an emerging yet challenging research area. In this paper, we deal with multimodal sarcasm and humor detection from conversational videos and image-text pairs. Being a fleeting action, which is reflected across the modalities, sarcasm detection is challenging since large datasets are not available for this task in the literature.
Therefore, we primarily focus on resource-constrained training, where the number of training samples is limited.
To this end, we propose a novel multimodal learning system, \textbf{\model} (\textbf{Mu}ltimodal \textbf{L}earning using \textbf{O}ptimal \textbf{T}ransport), which utilizes self-attention to exploit intra-modal correspondence and optimal transport for cross-modal correspondence. Finally, the modalities are combined with multimodal attention fusion to capture the inter-dependencies across modalities. 
We test our approach for multimodal sarcasm and humor detection on three benchmark datasets - MUStARD \cite{castro-etal-2019-towards} (video, audio, text), UR-FUNNY \cite{hasan-etal-2019-ur} (video, audio, text), MST \cite{cai-etal-2019-multi} (image, text) and obtain \textit{2.1\%}, \textit{1.54\%} and \textit{2.34\%} accuracy improvements over state-of-the-art.  

\end{abstract}

\section{Introduction}

With an abundance of user-generated multimodal content, such as videos, multimodal learning has become an important area of research \cite{wei2020multi, rahman2020integrating, atri2021see}. Unlike traditional unimodal learning on isolated modalities (such as vision, language, or acoustic), multimodal learning aims to aggregate complementary sources of information into a unified system. Understanding sarcasm from multimodal dialogues is a specialized form of sentiment analysis, where the speaker creatively experiments with words (language), gestures (vision), prosody (acoustic) to deliver incongruity across modalities. Humor is also a closely related quintessential sentiment often manifested using exaggeration or irony across modalities, such as a sudden twist of tone or a funny gesture. These two forms of subtle human sentiments are crucial to removing barriers in conversations, building trust \cite{vartabedian1993humor} and creating a positive impact on mental health \cite{lefcourt2012humor}. However, existing deep neural systems often struggle to understand such fine-grained multimodal sentiments.  

\begin{figure}[t]
\centering
\hspace*{0cm}
  \includegraphics[scale=0.090]{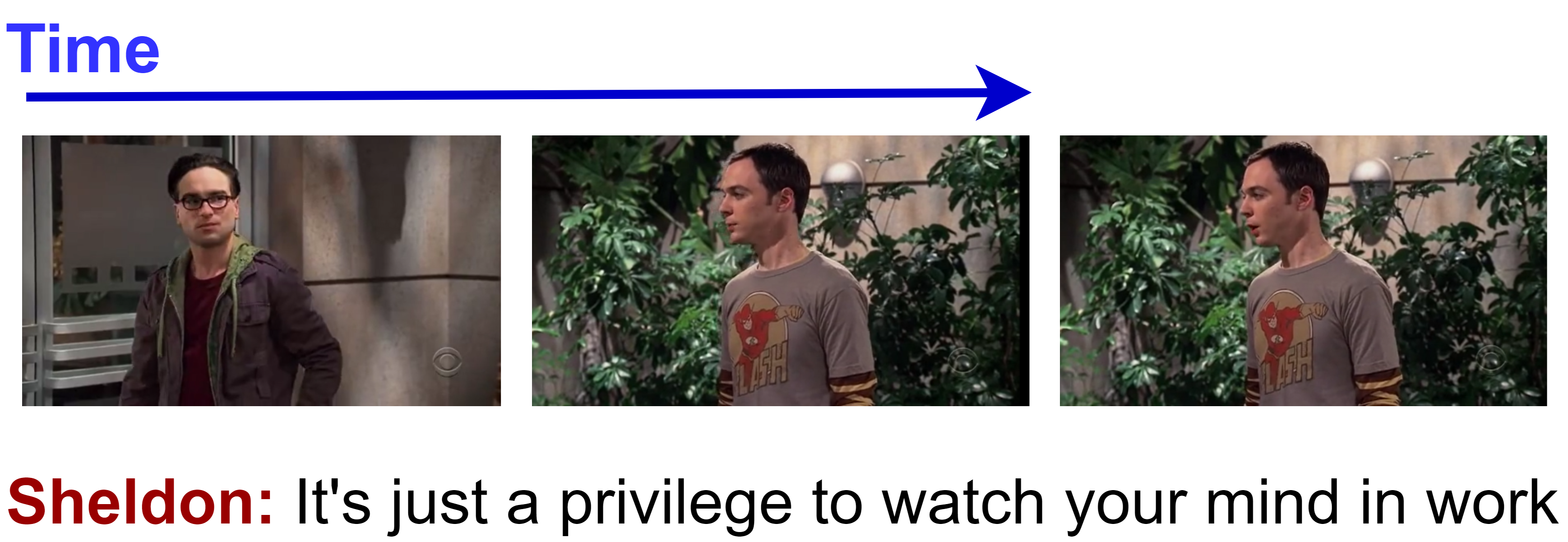}
  \caption{Sample sarcastic utterance from the MUStARD dataset along with its transcript. Sheldon's comment (text) with a straight face (visual) and a clinical tone (acoustic) makes the instance sarcastic.}
  \label{fig:mustard_example}
  \vspace{-0.5cm}
\end{figure}

\noindent \textbf{Motivation:} The most common form of sarcasm and humor has traditionally been delivered using text. However, sarcasm in multimodal data often requires precise inter-modal cues to reveal the speaker’s intentions.  For instance, it can often be expressed using a combination of verbal and non-verbal cues, such as a change of tone, overemphasis in a word, a drawn-out syllable, or a straight-looking face. Consider the example shown in Figure \ref{fig:mustard_example} - the seemingly applauding utterance \textit{"It's just a privilege to watch your mind in work"} becomes sarcastic when delivered with a straight face and clinical tone, overall expressing a negative connotation. Humans can naturally integrate all this simultaneous information subconsciously. However, building an algorithm that can potentially do the same requires appropriate representation of all these disparate sources of information and thus, gained immense research interest.

\noindent \textbf{Challenges:} Despite being regularly used in movies and sitcom shows, sarcastic conversational utterances are hard to collect, mainly because of the manual effort required in detecting and annotating raw videos with sarcasm labels. The only publicly available corpus, MUStARD, contains $690$ video samples with an even number of sarcastic and non-sarcastic labels. As a consequence, off-the-shelf multimodal transformer-based systems \cite{sun2019videobert, pan2020modeling, huang2020multimodal, rahman2020integrating} with a large number of parameters often tend to overfit this dataset. In this work, we propose \model, which utilizes multi-head self-attention and optimal transport (OT) - \cite{villani2008optimal} based domain adaptation to learn strong intra- and cross-modal dependencies using a small number of training samples. In order to embed asynchronous unimodal features in a common subspace, we use optimal transport kernel (OTK) Embedding \cite{mialon2021trainable}. Lastly, the self and cross-attended features are fused using an attention fusion module.

\noindent \textbf{Contributions:}  In summary, our contributions are three-fold. $(i)$ We propose a new multimodal learning method that uses multi-head self-attention and optimal transport (OT) - \cite{villani2008optimal} based domain adaptation to learn strong intra- and cross-modal dependencies using a small number of training samples. $(ii)$ We evaluate the proposed system on three different multimodal sarcasm and humor detection datasets and obtain $2.1-2.4\%$ improvements over the previous state-of-the-art. $(iii)$ We reduce the training corpus of two large multimodal datasets and demonstrate the superiority of our system over multimodal transformers in a resource-constrained training setup. On limited training samples, the proposed system outperforms state-of-the-art by an accuracy margin of $4.2-5.8\%$. 


\section{Related Work}
The proliferation of multimedia data on the Internet has resulted in multimodal systems gaining an increasing interest in recent years. One of the fundamental challenges in a multimodal framework is to fuse different unimodal features, which are asynchronous and have varying dimensions. Poria et al. \cite{poria2017review} presented an outline of various fusion techniques and potential performance improvements with multiple modalities.

\noindent \textbf{Multimodal Fusion:} Similar to human behavior, multimodal frameworks aim to integrate and correlate simultaneous multiple resources, such as acoustic, visual, and textual information \cite{castro-etal-2019-towards, hazarika2020misa, atri2021see, pramanick2021detecting}. There are mainly three types of fusion strategies – early, late, and hybrid. Early fusion methods \cite{poria2016convolutional, zadeh2017tensor} integrates different sources of data into a single feature vector and uses a single classifier. On the other hand, late fusion \cite{cambria2017affective, cao2016cross} aggregates the outputs of different classifiers trained on different modalities. However, none of these techniques consider the interdependence among the different modalities and hence perform poorly on real applications. In contrast, hybrid fusion \cite{zadeh2018memory, akhtar2019multi, wang2019words, pham2019found, pramanick2021exercise, pramanick2021momenta} jointly learns from various modality-specific sources by employing an intermediate shared representation layer and has been most prominent in the literature. Recently following the success of transformers \cite{vaswani2017attention, howard2018universal, peters2018deep, devlin-etal-2019-bert, yang2019xlnet, liu2019roberta}, several works have extended it to multimodal applications by capturing both unimodal and cross-modal interactions via fine-tuning \cite{tsai-etal-2019-multimodal, lu2019vilbert, sun2019videobert, pan2020modeling, huang2020multimodal, kant2020spatially, rahman2020integrating}. 
For example, Rahman et al. \cite{rahman2020integrating} integrate acoustic and visual information in pre-trained transformers like BERT \cite{devlin-etal-2019-bert} and XLNet \cite{yang2019xlnet}. However, all the multimodal transformers require a large number of training samples for good performance.  

\noindent \textbf{Multimodal Sarcasm \& Humor Detection:} Humor and sarcasm are two closely related sentiments - while humor is often expressed using exaggeration and irony, sarcasm mostly generates from incongruity. Unimodal humor and sarcasm detection are well-studied in literature \cite{tepperman2006yeah, woodland2011context, riloff2013sarcasm, yang2015humor, chandrasekaran2016we, liu2018modeling, chen2018humor, kolchinski2018representing}. However, the multimodal counterpart is more subtle as the irony or incongruity is often present in different modalities. Hasan et al. \cite{hasan-etal-2019-ur} introduced the first large-scale multimodal dataset (UR-FUNNY) for humor detection. Castro et al. \cite{castro-etal-2019-towards} collected a multimodal conversational dataset (MUStARD) for sarcasm detection from popular sitcom TV shows. Cai et al. \cite{cai-etal-2019-multi} developed another dataset to detect multimodal sarcasm in Twitter posts. Recently, Han et al. \cite{han2021bi} proposed an end-to-end network that performs fusion (relevance increment) and separation (difference increment) on pairwise modality representations for humor detection. Another closely related work, MISA \cite{hazarika2020misa} aggregates modality-invariant and modality-specific representations and has been applied to predict humor in the UR-FUNNY dataset. However, multimodal sarcastic video samples are hard to collect, and the existing neural systems struggle to perform well on MUStARD. In this paper, we utilize self-attention to capture intra-modal correspondence and optimal transport for cross-modal correspondence in a low-resource training setup.

\section{\model: Proposed System}
This section describes our proposed system, \model\, for sarcasm and humor detection in videos and images by leveraging multimodal signals. Each video consists of multiple utterances\footnote{An utterance is a unit of speech bounded by breaths or pauses \cite{olson1977utterance}}, where each utterance - a smaller video by itself, is considered to contain sarcastic/humorous dialogues for positive samples. These utterances are a multimodal source of data - we preprocess the input utterances to extract fine-grained unimodal features from three modalities - visual $(v)$, language $(l)$, and acoustic $(a)$. Additionally, every utterance is accompanied by a context - which is also a short video helping to discern the background of the utterance. We concatenate the context with the utterance so that the utterance can be modeled in light of the context and the resulting video is treated as the input to the system.  Consequently, our system is generalizable to images. Each input image is associated with a caption and a piece of text on it\footnote{We employ Google OCR Vision API to extract this text}, which we consider as visual $(v)$, caption text $(t_c)$ and OCR text $(t_{ocr})$ modalities. Features corresponding to these modalities are represented as $U_m \in \mathbb{R}^{L_m \times d_m}$, where $L_m$ denotes the sequence length for modality $m$ and $d_m$ is the respective feature dimension. The details of these features are discussed in Section \ref{sec:feature_extraction}. Given these unimodal feature sequences $U_{m \in \{v, l/t_c, a/t_{ocr}\}}$, the task is to predict the affective orientation of the input in a binary classification setup - sarcastic/humorous or non-sarcastic/non-humorous. 

\begin{figure*}[t!]
\centering
\hspace*{0cm}
  \includegraphics[scale=0.058]{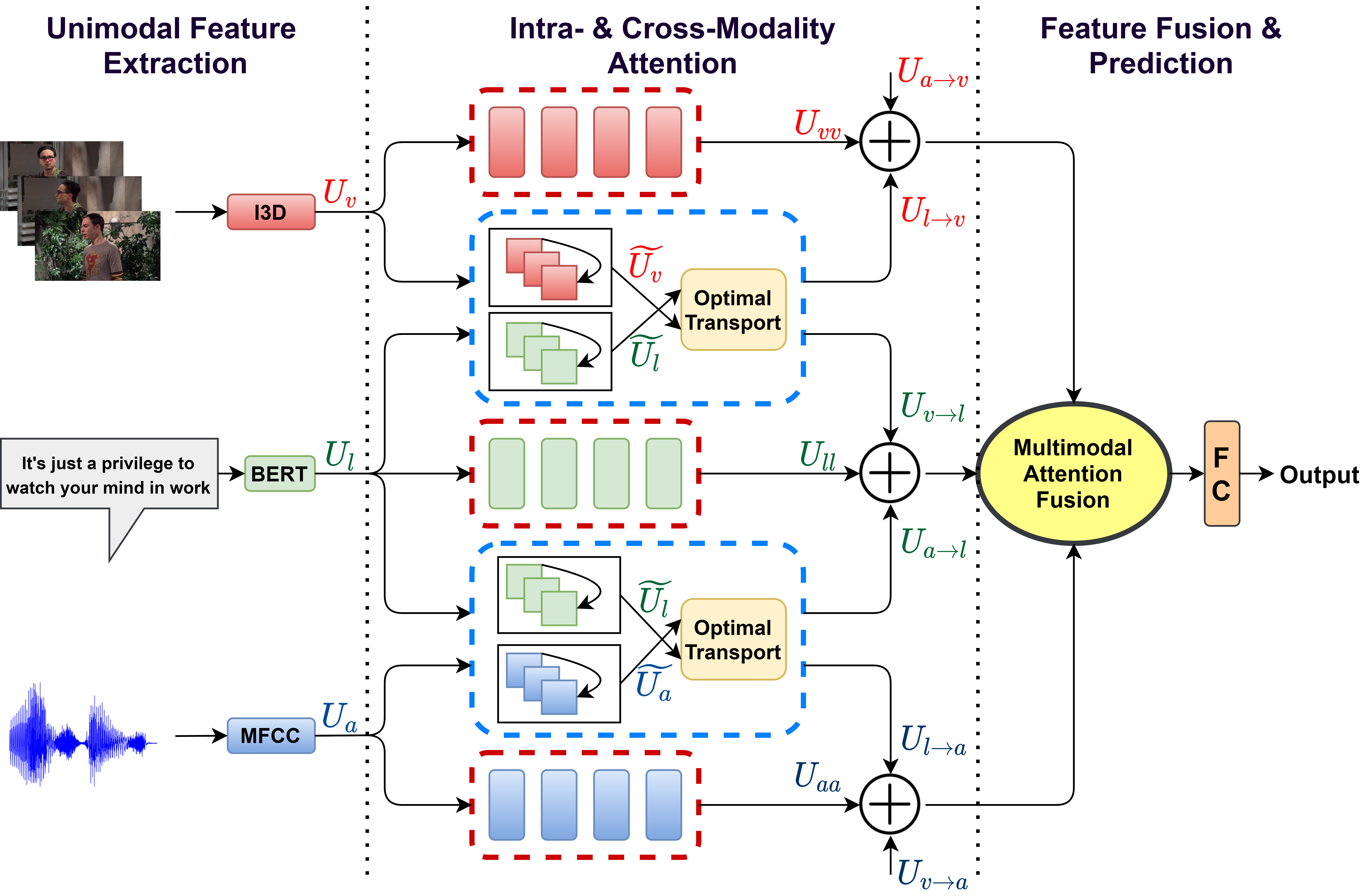}
  \vskip-5pt\caption{The architecture of the proposed model, \model. The features from each modality are passed through multi-head self-attention module (\textcolor{red}{red dashed}) and optimal transport-based cross-attention module (\textcolor{blue}{blue dashed}). Finally, the attended features are fused using Multimodal Attention Fusion (MAF).}
  \label{fig:model}
  \vspace{-0.5cm}
\end{figure*}

As shown in Figure \ref{fig:model},  \model\ consists of two modules, the self-attention module and the cross-attention module, shown in red and blue dashed blocks, respectively. First, the modality-specific fine-grained features are fed into the self-attention module which aims to learn intra-modality correspondence by emphasizing the most relevant tokens in the entire feature sequence. In parallel, the cross-attention module maps pairwise unimodal modalities into a common feature space, and thus learns inter-modality interaction.  By taking both intra-modality and inter-modality relationships into consideration, the system gains the ability to synchronize across different modalities over various time and learns the correlation among them \cite{wei2020multi}. Finally, the self-attended unimodal features and cross-attended shared features are fused by using an attention fusion mechanism. It is important to observe that \model\ achieves the same objective as multimodal transformers \cite{tsai-etal-2019-multimodal, lu2019vilbert, sun2019videobert, pan2020modeling, huang2020multimodal, kant2020spatially, rahman2020integrating}, but has significantly fewer trainable parameters, and thus works better in low-resource training setup.  

\subsection{Unimodal Feature Extraction} \label{sec:feature_extraction}
In the proposed multimodal learning setup, we leverage three modalities, e.g., video, audio and text. Initially we extract fine-grained unimodal features from these modalities as described below.

\noindent \textbf{Visual Features:} The video features are extracted using a pre-trained action recognition model, $I3D$ \cite{carreira2017quo} trained on the Kinetics dataset \cite{kay2017kinetics} to recognize $400$ different human actions. All the frames, computed at a rate of $5$ FPS, are first preprocessed by resizing, center-cropping, and normalization to have a resolution of $112 \times 112$. For every $16$ non-overlapping frames in a video, $I3D$ extracts a $2048$ dimensional feature vector. Therefore, the final unimodal feature dimension for every video is $L_v \times 2048$, where $L_v$ is the number of sets $16$ non-overlapping frames. For images, we employ ResNet-101 \cite{he2016deep} pre-trained on ImageNet \cite{deng2009imagenet}. Specifically, we extract $7\times 7\times 2048$ feature maps from the last pooling layer (pool5) of ResNet-101 and reshape it into a dimension of $49 \times 2048$. Hence, for the images, $L_v = 49$.

\noindent \textbf{Language Features:} Traditionally, language modality features have been GloVe \cite{pennington2014glove} or Word2Vec \cite{mikolov2013efficient} embeddings for each token. However, following recent works \cite{castro-etal-2019-towards, sun2020learning, hazarika2020misa, pan2020modeling}, we utilize the pre-trained BERT \cite{devlin-etal-2019-bert} as the feature extractor for textual utterances. In particular, we input the raw text to a pre-trained uncased BERT-base model to get a $768$-dimensional dynamic feature representation for every token in the utterance. Hence, the resulting language feature dimension is $L_l \times 768$, where $L_l$ is the number of tokens in the utterance. 

\noindent \textbf{Acoustic Features:} The acoustic modality is expected to contribute information related to pitch, intonation, and other tonal-specific details of the speaker \cite{tepperman2006yeah}. To achieve this, we obtain low-level features from the audio stream for each video. Following state-of-the-art, we use COVAREP \cite{degottex2014covarep} to extract the audio features. These low-level features include but not limited to $12$ Mel-frequency cepstral coefficients, Voiced/Unvoiced segmenting features (VUV) \cite{drugman2011joint}, glottal source parameters \cite{drugman2011detection}\footnote{For the full set of features, please refer to \cite{zadeh2018multimodal, hasan-etal-2019-ur}}. The resulting acoustic feature dimension is $L_a \times 81$, where $L_a$ depends on the length of the video. Note that we use the OT kernel module in our system to align all three temporal sequences within an utterance to be of equal length, i.e. $L_v = L_l = L_a$.

\subsection{Intra-Modality Attention} Now we introduce the self-attention module used to model the intra-modal correspondence among three different modalities. Typically, an attention module can be described as a tuple of key, query and value, where the output is a weighted sum of the values and the weight matrix is determined by query and its corresponding key. In case of self-attention, the key, query and value are equal. Following the philosophy of \cite{vaswani2017attention}, the recent trend is to apply transformers for modeling self-attention. However, a transformer module consists of multiple sub-layers and is often difficult to train from scratch. In our low-resource training setup, we simplify the structure of a transformer module by utilizing the multi-head attention layers.  

We compute the multi-head self-attention as follows.  Given the unimodal feature matrices $U_m \in \mathbb{R}^{L_m \times d_m}$, where $m \in \{v, l/t_c, a/t_{ocr}\}$, we aim to extract $k$ distinct set of relevant features corresponding to the input representation, where $k$ is a hyper-parameter denoting the number of attention heads. The multi-head attention mechanism takes $U_m$ as an input, and outputs an attention weight matrix $W_m \in \mathbb{R}^{k\times L_m}$, respective to the modality-specific representation as follows
\setlength{\belowdisplayskip}{0pt} \setlength{\belowdisplayshortskip}{0pt}
\setlength{\abovedisplayskip}{0pt} \setlength{\abovedisplayshortskip}{0pt}
\begin{equation} \label{equ:weight_self_attention}
    W_m = softmax(W_{h2} \tanh(W_{h1} U_m^\top)),
\end{equation}
where, ${W_{h1}} \in \mathbb{R}^{r \times d_m}$ and  ${W_{h2}} \in \mathbb{R}^{k\times r}$ are parameter matrices to be learned during training. The $softmax(\cdot)$ is performed along the second dimension of its input, and $r$ is a hyper-parameter we can set arbitrarily.  The resulting modality-specific embedding matrices $U_{mm}$ are computed using their respective attention weights and unattended features as follows
\begin{equation}
    U_{mm} = W_m U_m,
\end{equation}
where, $U_{mm}\in \mathbb{R}^{k\times d_m}$ are the resulting self-attended features for modality $m$. 

\subsection{Cross-Modality Attention} Although the self-attention module effectively exploits the intra-modality correspondence, the inter-modality relationship, i.e. the interdependence across three different modalities is not deployed. Motivated by the success of Transformers \cite{vaswani2017attention, howard2018universal, peters2018deep, devlin-etal-2019-bert, yang2019xlnet, liu2019roberta}, several recent works have extended it to learn multimodal association by adding additional vision and speech modules to the Transformer framework \cite{tsai-etal-2019-multimodal, lu2019vilbert, sun2019videobert, pan2020modeling, huang2020multimodal, kant2020spatially, rahman2020integrating}. Although multimodal transformers rely on a language-pretrained BERT \cite{devlin-etal-2019-bert}, which is fixed throughout, the vision and acoustic components have to be trained from scratch, which often requires hundreds of thousands of training samples. To this end, instead of using a transformer-based cross-attention module, we exploit a recently proposed technique viz., optimal transport kernel (OTK) \cite{mialon2021trainable} for modelling the inter-modality association. OTK combines the idea of kernel methods and optimal transport to fuse the unimodal features with varying dimension and dependencies into a same reference frame. The complete cross-attention module consists of two steps --  unimodal feature transform and domain adaptation using optimal transport.

\begin{table*}[t!]
\centering
    
\resizebox{0.95\textwidth}{!}
{
\begin{tabular}{c | c || c c c || c c c || c c c}

\multirow{2}{*}{Dataset} & \multirow{2}{*}{\#Samples} & \multicolumn{3}{c}{Train} & \multicolumn{3}{c}{Dev} & \multicolumn{3}{c}{Test}\\ \cline{3-11}
& & \#Sar & \#Non-Sar & Total & \#Sar & \#Non-Sar & Total & \#Sar & \#Non-Sar & Total \\ \hline

\textbf{MUStARD} & 690 & 275 & 275 & 550 & 35 & 35 & 70 & 35 & 35 & 70 \\ 
\textbf{MST} & 24635 & 8642 & 11174 & 19816 & 959 & 1451 & 2410 & 959 & 1450 & 2409 \\
\textbf{Tiny MST} & 7819 & 2000 & 2000 & 4000 & 959 & 1451 & 2410 & 959 & 1450 & 2409 \\ \hline \hline

& & \#Hum & \#Non-Hum & Total & \#Hum & \#Non-Hum & Total & \#Hum & \#Non-Hum & Total \\ \cline{3-11}
\textbf{UR-FUNNY} & 16514 & 5306 & 5292 & 10598 & 1313 & 1313 & 2626 & 1638 & 1652 & 3290 \\
\textbf{Tiny UR-F} & 8916 & 1500 & 1500 & 3000 & 1313 & 1313 & 2626 & 1638 & 1652 & 3290 \\

\hline

\end{tabular}}

\caption{Statistics about the five datasets used in our experiments. We have reduced the training set of MST and UR-FUNNY to generate Tiny MST and Tiny UR-F, respectively. However, the corresponding dev and test sets remain unchanged.}
\vspace{-0.3cm}
\label{tab:dataset}
\end{table*}

\noindent {\bf{Unimodal Feature Transform:}} Since the visual,  language and acoustic features have different dimensions with varying sizes and dependencies, mapping those into a same reference frame is necessary yet challenging.  In the OTK approach, the unimodal feature vectors are embedded into a reproducing kernel Hilbert  space  (RKHS)  and  then  a weighted pooling operation is performed based on attention with weights provided by the transport plan between the set and a trainable reference.  The primary motivation for using kernels is to provide non-linear transformation of the input features. Afterward, optimal transport aligns the features on a trainable reference frame.

Suppose we want to compute an optimal transport plan from $x$ to $x'$. Let $a$ and $b$ be defined as $a = \sum_{i} a_i\delta_{i}$ and $b = \sum_{i} b_i\delta_{i}$ and $C$ be the pairwise costs for aligning the elements $x$ to $x'$. Then the entropic regularized
Kantorovich relaxation of OT from $x$ to $x'$ is~\cite{villani2008optimal}:
\begin{equation}
    \min_{P \in U(a,b)} \sum_{ij} C_{ij} P_{ij} - \epsilon H(P),
\label{eq:OT}
\end{equation}
where $H(P) = - \sum_{ij} P_{ij}(\log(P_{ij}-1))$, is the entropic regularization with parameter $\epsilon$, which controls the sparsity of $P$ and $U$ is the subspace of admissible coupling between $a$ and $b$ defined as follows
\begin{equation}
    U(a,b) = \{ P \in R_{+}: P1_{n} = a \text{ and } P^T1_{n'} = b   \}.
\end{equation}
Now, suppose $x = (x_1, x_2, .., x_n)$ be input feature vector and $z = (z_1, z_2, .., z_k)$ be the reference set with $k$ elements. Let $\kappa$ be a positive definite kernel and $\phi$ its corroesponding kernel embedding. 
The $\kappa$ matrix contains the values of $\kappa(x_i,z_i)$ before alignment and the transportation plan between $x$ and $z$ is denoted as $P(x,z)$, which is defined by the unique solution of Eq.~\eqref{eq:OT} when choosing cost $C = -\kappa$.
The embedding is defined as:
$\Phi_{z}(x) = \sqrt{p} P(x,z)^T\phi(x)$,
where $\phi(x) = [\phi(x_1), .., \phi(x_n)]^T$.  $\Phi_{z}(x)$ can also be represented as a positive semidefinite kernel as follows~\cite{mialon2021trainable}
\begin{equation}
\nonumber K_z(x, x') = \sum P_z(x,x')_{ij} \kappa(x_{i}, x_{j}) = \langle \Phi_{z}(x), \Phi_{z}(x'), \rangle 
\end{equation}
where, $K_z$ is the OTK. Using OTK, we transform $L_m \times d_m$ dimensional unimodal features to a uniform reference frame with length $L_{uni}$ and feature dimension $d_{uni}$, where $L_{uni}$ and $d_{uni}$ are chosen by experiments. The transformed features are represented as $\widetilde{U_m} \in \mathbb{R}^{L_{uni} \times d_{uni}}$, where $m \in \{v, l/t_c, a/t_{ocr}\}$.

\noindent \textbf{Domain Adaptation using Optimal Transport:} After transforming the varying dimensional unimodal features into a uniform reference frame, we finally embed pairwise unimodal features into a common subspace, using the OT-based domain adaptation. To implement OT as part of a larger learning model, we use the recently released Python optimal transport library \cite{flamary2021pot} in our experiments. 

We transport among each pair of modalities, which can be interpreted as domain adaptation across two modalities and the transported features are then concatenated to obtain the final adapted features. $U_{n \to m}$ denotes the transported features from modality $n$ to modality $m$, i.e.
\begin{equation}
    U_{n \to m} = OT(\widetilde{U_n} \to \widetilde{U_m}).
\end{equation}
We then concatenate transported features to the self-attended features $U_{mm}$ to obtain the final adopted features as follows
\begin{equation}
    U^m_{shared} = U_{mm} \oplus U_{n \to m} \oplus U_{p \to m},
\end{equation}
where, $m,n,p \in \{v, l/t_c, a/t_{ocr}\}$, $m \neq n \neq p$, and $\oplus$ denotes simple concatenation.

\subsection{Multimodal Attention Fusion (MAF)}

MAF utilizes an attention-based mechanism to fuse three set of features, $U^m_{shared}$, corresponding to the three modalities. For some samples, the visual modality is relevant, while for others, the language or acoustic modality plays more crucial role. Hence, the MAF module aims to model the relative important of different modalities. Motivated by \cite{gu2018hybrid, pramanick2021exercise}, we design our MAF module with two major parts -- modality attention generation and weighted feature concatenation. In the first part, a sequence of dense layers followed by a softmax layer is used to generate the attention scores $[w_v, w_l, w_a]$ for the three modalities.  In the second part, the adopted features are weighted using their respective attention scores and concatenated together as follows 
\begin{equation}
    U^{v}_{final} = (1 + w_v)U^v_{shared}
\end{equation}
\vspace{-0.22cm}
\begin{equation}
    U^{l}_{final} = (1 + w_l)U^l_{shared}
\end{equation}
\vspace{-0.2cm}
\begin{equation}
    U^{a}_{final} = (1 + w_a)U^a_{shared}
\end{equation}
\vspace{-0.2cm}
\begin{equation}
    U^{v,l,a}_{final} = {W_U}\otimes [U^{v}_{final}, U^{l}_{final}, U^{a}_{final}]
\end{equation}
We also use residual connections for better gradient flow. 
The final multimodal representation, $U^{v,l,a}_{final}$, is fed into a series of fully-connected layers for the final binary classification.

\section{Experiments}

In this section, we present details of the datasets used, unimodal feature extraction steps, baselines, and training methodologies.

\subsection{Datasets}
We evaluate \model on three different benchmark datasets for multimodal sarcasm and humor detection. Additionally, we access the effectiveness of \model\ in a low-resource setup by shrinking the large training sets of two datasets. 

\noindent \textbf{MUStARD:} Multimodal Sarcasm Detection Dataset (MUStARD) \cite{castro-etal-2019-towards} is the only available resource to enable sarcasm detection in conversational videos. This dataset is curated from popular TV shows like Friends, The Big Bang Theory and consists of audiovisual utterances annotated with sarcasm labels. Each target utterance in this dataset is associated with historical dialogues as context, which is key to understanding the backdrop of sarcastic remarks. The primary challenge in using MUStARD is its small size, which limits the performance of heavy transformer-based systems on this corpus.   

\noindent \textbf{MST:} Multimodal Sarcasm in Twitter Posts (MST) \cite{cai-etal-2019-multi} consists of sarcastic and non-sarcastic image-text pairs collected from Twitter. This dataset is primarily bimodal, as there is no acoustic modality. However, we notice that a significant amount of samples contains textual information on the image. We employ Google OCR Vision API\footnote{\url{https://cloud.google.com/vision/docs/ocr}} to extract this text and treat this as the third modality. We further reduce the training set of MST and name the resulting corpus Tiny MST.  

\noindent \textbf{UR-FUNNY:} For multimodal humor detection, we use the UR-FUNNY dataset \cite{hasan-etal-2019-ur} which is collected from TED talk videos and therefore has three modalities. Similar to MUStARD, this dataset consists of context preceding
the target punchline. We further create a tiny variant of UR-FUNNY by truncating its training set. The split statistics of all these datasets are presented in Table \ref{tab:dataset}.

\subsection{Baselines}
For unified comparison across videos and images, we use the following baselines.
Furthermore, we remove various modalities and modules at a time from the proposed \model\ system to observe the effect in performance. Baselines on MUStARD and UR-FUNNY include -
\textbf{Support Vector Machines (SVM)}~\cite{cortes1995support}, \textbf{DFF-ATMF} \cite{chen2019complementary},
\textbf{CIM-MTL} \cite{akhtar2019multi},
\textbf{Tensor Fusion Network (TFN)} \cite{zadeh2017tensor},
\textbf{Contextual Memory Fusion Network (CMFN)} \cite{hasan-etal-2019-ur},
\textbf{MISA} \cite{hazarika2020misa},
\textbf{MAG-XLNet} \cite{rahman2020integrating}. MAG-XLNet introduced Multimodal Adaption Gate (MAG) to fuse acoustic and visual information in pre-trained language transformers, and is the SOTA on both MUStARD and UR-FUNNY.  Baselines on MST include-
\textbf{Concat BERT} (concatenates the features extracted by pre-trained unimodal ResNet-152 \cite{he2016deep} and Text BERT \cite{devlin-etal-2019-bert} and uses a simple perceptron as the classifier), 
\textbf{Supervised Multimodal Bitransformer (MMBT)} \cite{kiela2020supervised},
\textbf{Vision and Language BERT (ViLBERT)} \cite{lu2019vilbert},
\textbf{Hierarchical Fusion Model (HFM)} \cite{cai-etal-2019-multi},
\textbf{D\&R Net} \cite{xu2020reasoning}, 
\textbf{MsdBERT} \cite{pan2020modeling}.
\textbf{MsdBERT}, which exploits a co-attention network to exploit intra- and inter-modality incongruity between text and image, is the SOTA method on the MST dataset.   
More details on the baselines are provided in supplementary document.

\subsection{Training}

We train \model\ using Pytorch framework \cite{paszke2019pytorch} on Nvidia-RTX 2080Ti GPUs, with 24 GB dedicated memory in each GPU. As described in Section \ref{sec:feature_extraction}, we use pre-trained unimodal feature extraction models, and fine-tune them during training. All other weights are randomly initialized with a zero-mean Gaussian distribution with standard deviation 0.02.

Although MUstARD and UR-FUNNY are balanced datasets, 
Table \ref{tab:dataset} shows label imbalance for the MST dataset. Therefore, when training on MST dataset, we assign larger weight ($[w_{sar},w_{non-sar}]=[1.2, 1]$) for minority class to minimize label imbalance. We train our model using the Adam \cite{loshchilov2018decoupled} optimizer and the binary cross-entropy loss as the objective function. The initial learning rate is $0.005$ and the network is trained for $300$ epoches. The detailed hyper-parameters used for the training is provided in the supplementary material.

\section{Results, Discussion and Analysis}

In this section, we compare the performance of \model\ system with different multimodal baselines, conduct a detailed ablation study to demonstrate the importance of different modalities and modules in our system. Furthermore, we visualize the interpretability of \model\ using Grad-CAM \cite{selvaraju2017grad}. Since the test sets of MUStARD and UR-FUNNY are balanced, we use binary accuracy as the evaluation metric. However, since the test set of MST is imbalanced, we also consider F1 scores when evaluating on this dataset.

\begin{table}[t]
\centering 
\resizebox{1\columnwidth}{!}
{
\begin{tabular}{l | c c || c | c | c}
\multirow{2}{*}{\bf Algorithm} & \multirow{2}{*}{\bf Context} & \multirow{2}{*}{\bf Target} &  \multicolumn{1}{c|}{\bf MUStARD} & \multicolumn{1}{c|}{\bf UR-FUNNY} & \multicolumn{1}{c}{\bf Tiny UR-F} \\

& & & Acc $\uparrow$ & Acc $\uparrow$ & Acc $\uparrow$  \\

\hline

SVM & \xmark & \cmark & 73.55 & - & - \\
DFF-ATMF & \xmark & \cmark & 64.45 & 62.55 & 56.35\\
CIM-MTL & \xmark & \cmark & 67.14 & 63.20 & 56.71\\
TFN & \xmark & \cmark & 68.57 & 64.71 & 57.23\\
\hline 
CMFN (GloVe) & \xmark & \cmark & 67.14 & 64.47 & 57.10\\
CMFN (GloVe) & \cmark & \cmark & 70.00 & 65.23 & 59.25\\
\hline 
MISA (BERT) & \xmark & \cmark & 66.18 & 70.61 & 62.66\\
BBFN & \cmark & \cmark & 71.42 & 71.68 & 63.20\\
MAG-XLNet & \cmark & \cmark & 74.72 & 72.43 & 67.22\\
\hline 
\hline
\model & \xmark & \cmark & 74.52 & 73.22 & 70.74\\
\model & \cmark & \cmark & \bf 76.82$^\dagger$ & \bf 73.97$^\dagger$ & \bf 71.46$^\dagger$\\

\hline

\multicolumn{3}{c||}{\bf {$\Delta_{\model-baseline}$}} & \textcolor{blue}{$\uparrow$ 2.10} & \textcolor{blue}{$\uparrow$ 1.54} & \textcolor{blue}{$\uparrow$ 4.24} \\

\hline
\end{tabular}}
\vspace{-0.2cm}
\caption{Performances of multimodal models on the MUStARD, UR-FUNNY and Tiny UR-F datasets. Since the test sets are balanced (c.f. Table \ref{tab:dataset}), only binary classification accuracy is reported. $^\dagger$ indicates the results have experienced paired t-test with $p < 0.01$ and demonstrate significant improvement over MAG-XLNet, the best baseline model.}
\label{tab:results_video}
\end{table}

\begin{table}[t]
\centering 
\resizebox{0.95\columnwidth}{!}
{
\begin{tabular}{l | c || c c | c c}
\multirow{2}{*}{\bf Algorithm} & \multirow{2}{*}{\bf OCR} &  \multicolumn{2}{c|}{\bf MST} & \multicolumn{2}{c}{\bf Tiny MST} \\

& & Acc $\uparrow$ & F1 $\uparrow$ & Acc $\uparrow$ & F1 $\uparrow$  \\

\hline

Concat BERT & \xmark & 81.08 & 79.56 & 76.21 & 73.48\\
HFM & \xmark & 83.44 & 80.18 & 77.80 & 74.07\\
D\&R Net & \xmark & 84.02 & 80.60 & 79.43 & 76.72\\
\hline
MMBT & \xmark & 83.46 & 80.74 & 79.48 & 76.09\\
MMBT & \cmark & 84.87 & 82.66 & 80.57 & 77.20\\
\hline
ViLBERT & \xmark & 84.21 & 82.49 & 79.42 & 75.95\\
ViLBERT & \cmark & 86.90 & 84.22 & 80.68 & 77.24\\
\hline
MsdBERT & \xmark & 86.05 & 82.92 & 80.14 & 77.53\\
MsdBERT & \cmark & 88.75 & 86.18 & 82.30 & 79.90\\
\hline 
\hline
\model & \xmark & 87.41 & 86.33 & 84.46 & 82.62\\
\model & \cmark & \textbf{90.82$^\dagger$} & \textbf{88.52$^\dagger$} &\textbf{88.04$^\dagger$} & \textbf{85.93$^\dagger$}\\

\hline
\multicolumn{2}{c||}{\bf {$\Delta_{\model-baseline}$}} & \textcolor{blue}{$\uparrow$ 2.07} & \textcolor{blue}{$\uparrow$ 2.34} & \textcolor{blue}{$\uparrow$ 5.74} & \textcolor{blue}{$\uparrow$ 6.03}\\
\hline

\end{tabular}}
\caption{Performance of multimodal models on the MST and Tiny MST datasets. Since the test set is imbalanced (c.f. Table \ref{tab:dataset}), both binary classification accuracy and macro F1 scores are reported. $^\dagger$ indicates the results have experienced paired t-test with $p < 0.01$ and demonstrate significant improvement over MsdBERT, the state-of-the-art model.}
\label{tab:results_image}
\vspace{-0.5cm}
\end{table}

\begin{table*}
\centering 
\resizebox{1.9\columnwidth}{!}
{
\begin{tabular}{c | l || c | c | c || l || c c | c c}
\multirow{2}{*}{\bf Modality} & \multirow{2}{*}{\bf Algorithm} &  \multicolumn{1}{c|}{\bf MUStARD} & \multicolumn{1}{c|}{\bf UR-FUNNY} & \multicolumn{1}{c||}{\bf Tiny UR-F} & \multirow{2}{*}{\bf Algorithm} & \multicolumn{2}{c|}{\bf MST} & \multicolumn{2}{c}{\bf Tiny MST} \\

& & Acc $\uparrow$ & Acc $\uparrow$ & Acc $\uparrow$ & & Acc $\uparrow$ & F1 $\uparrow$ & Acc $\uparrow$ & F1 $\uparrow$ \\

\hline

\tt Trimodal & \bf \model & \bf 78.57 & \bf 73.97 & \bf 71.46 & \bf \model & \bf 90.82 & \bf 88.52 & \bf 88.04 & \bf 85.93 \\ 

\hline

\multirow{3}{*}{\tt Unimodal} & \texttt{visual} only & 73.30 & 60.72 & 58.80 & \texttt{visual} only & 82.65 & 81.22 & 78.56 & 77.70 \\
& \texttt{language} only & 73.54 & 69.58 & 67.32 & \texttt{caption} only & 83.40 & 82.14 & 80.06 & 78.85 \\
& \texttt{acoustic} only & 64.00 & 64.35 & 55.44 & \texttt{OCR} only & 78.64 & 77.39 & 76.22 & 75.31 \\

\hline

\multirow{3}{*}{\tt Bimodal} & \texttt{visual} + \texttt{language} & 77.18 & 70.40 & 69.40 & \texttt{visual} + \texttt{caption} & 87.35 & 85.93 & 83.94 & 82.45 \\
& \texttt{visual} + \texttt{acoustic} & 75.54 & 69.23 & 69.82 & \texttt{visual} + \texttt{OCR} & 85.66 & 84.37 & 82.30 & 81.38 \\
& \texttt{language} + \texttt{acoustic} & 75.72 & 72.10 & 69.12 & \texttt{caption} + \texttt{OCR} & 85.10 & 83.79 & 81.87 & 81.00 \\

\hline 

\multirow{3}{*}{\tt Trimodal} & \model\ w/o self-att$^n$ & 71.60 & 64.46 & 63.22 & \model\ w/o self-att$^n$ & 86.24 & 84.80 & 83.69 & 82.84 \\
& \model\ w/o cross-att$^n$ & 63.88 & 60.08 & 57.15 & \model\ w/o cross-att$^n$ & 82.32 & 80.20 & 80.07 & 79.28 \\
& \model\ w/o MAF & 75.23 & 71.22 & 68.84 & \model\ w/o MAF & 87.94 & 86.73 & 85.55 & 84.38 \\

\hline

\end{tabular}}
\caption{Ablation Study. Role of different modalities and various modules in our proposed \model\ system on five datasets. }
\label{tab:results_ablation}
\vspace{-0.5cm}
\end{table*}

\subsection{Comparison with Baselines}
\noindent \textbf{Multimodal Sarcasm Detection:} Table \ref{tab:results_video} presents the comparative classification performances on the MUStARD dataset. Since this corpus is relatively small, complex neural models with a large number of parameters often overfit when trained from scratch. The authors of the MUStARD paper reported the best accuracy of $71.60\%$ when using a simple SVM as classifier \cite{castro-etal-2019-towards}. However, they used framewise ResNet features as the visual modality, which does not consider the temporal dynamics of the videos. We retrain an SVM using I3D features and produce $73.55\%$ accuracy to indicate the effectiveness of I3D features for videos. Transformer-based inter-utterance contextual baselines, DFF-ATMF and CIM-MTL, perform poorly on this dataset. We also observe the role of historical context dialogues to model sarcasm. The same CMFN baseline yields $2.86\%$ better accuracy after using context. Nevertheless, baselines like MISA and TFN do not improve from context. As described by the respective authors, these models suffer in encoding the long sequence and thus, lose crucial information during fusion.

MAG-XLNet, which is the current state-of-the-art model in CMU-MOSI and CMU-MOSEI datasets for multimodal sentiment analysis, also performs well on MUStARD. Unlike other multimodal transformers, MAG-XLNet makes no change to the original structure of XLNet, but rather comes as an attachment to utilize multimodal information and hence proves relatively more effectual in a low resource setup. Our proposed system, \model, improves on MAG-XLNet by $2.10\%$. It is important to note that, \model\ has only $21$M trainable parameters, which is almost $8$ times smaller than MAG-XLNet. Domain adaptation using OT and low-number of parameters helps \model\ in achieving the best results. 

Table \ref{tab:results_image} shows the performance of baselines and \model\ on the MST dataset. Since a significant amount of the image samples of this dataset have text on it, the state-of-the-art system, MsdBERT utilizes OCR extracted texts to improve its F$1$ score by $3.26\%$. MMBT and ViLBERT also show a very similar trend. Since the MST dataset has a large training corpus, sophisticated transformer-based systems significantly outperform simple models like Concat BERT and HFM. Our proposed system, \model\ beats the SOTA on MST by an F1 score of $2.34\%$, demonstrating its effectiveness on large corpus as well. Furthermore, in order to evaluate \model\ on a low amount of training data, we randomly reduce the training set of MST to have only $2000$ samples in each class and call it Tiny MST. When trained on Tiny MST, the performance of MMBT, ViLBERT, and MsdBERT drops drastically. However, \model\ is able to maintain an F1 score of $85.93\%$ and beats the best baseline by a substantial margin of $6.03\%$. The low number of trainable parameters and optimal transport-based domain adaptation setup in \model\ again proves helpful when trained with limited data. 

\noindent \textbf{Multimodal Humor Detection:} For humor detection, we present the classification performances of baselines and \model\ in Table \ref{tab:results_video}. Similar to MUStARD, MAG-XLNet produces the best baseline performance on the UR-FUNNY dataset. The inclusion of context also helps many baselines to boost their performance by a few points. \model\ beats MAG-XLNet marginally on UR-FUNNY. However, when the training corpus size is reduced, MAG-XLNet drops its performance by $5.21\%$ compared to \model's $2.51\%$. Thus, on all three low-resource setups, \model\ consistently outperforms all the baselines by a large margin, demonstrating that multimodal transformers are not good enough in the absence of ample training samples.

\begin{figure*}[ht!]
\centering
\subfloat{
\includegraphics[width=0.21\textwidth, height=0.16\textwidth]{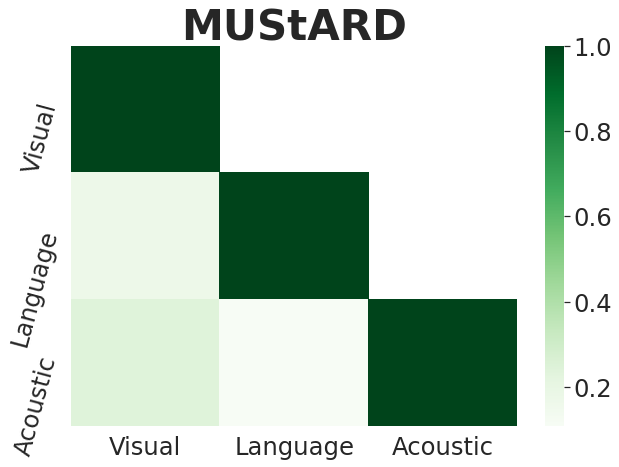}}\hspace{0.1em}
\subfloat{
\includegraphics[width=0.178\textwidth, height=0.16\textwidth]{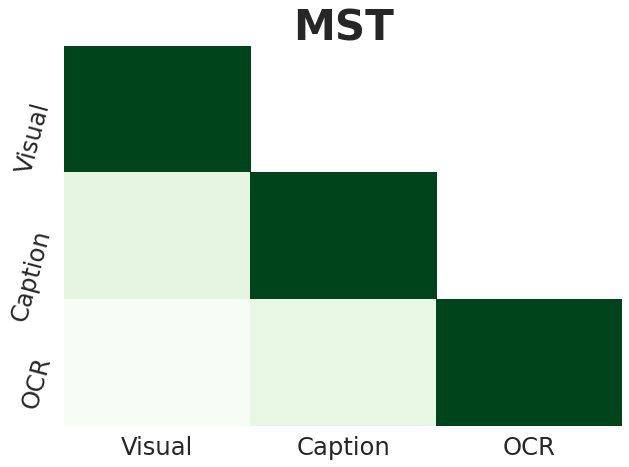}}
\subfloat{
\includegraphics[width=0.178\textwidth, height=0.16\textwidth]{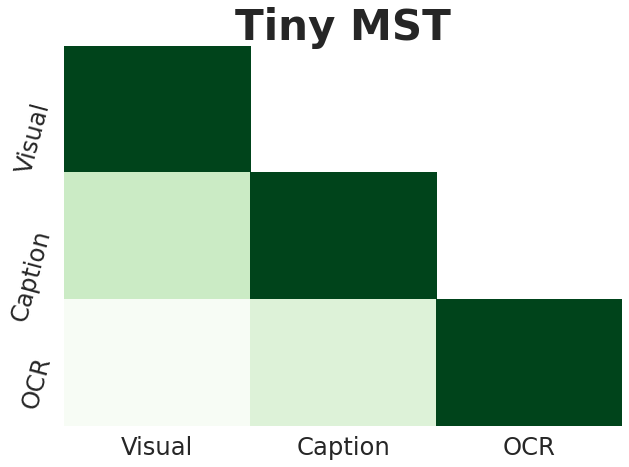}}\hspace{0.1em}
\subfloat{
\includegraphics[width=0.178\textwidth, height=0.16\textwidth]{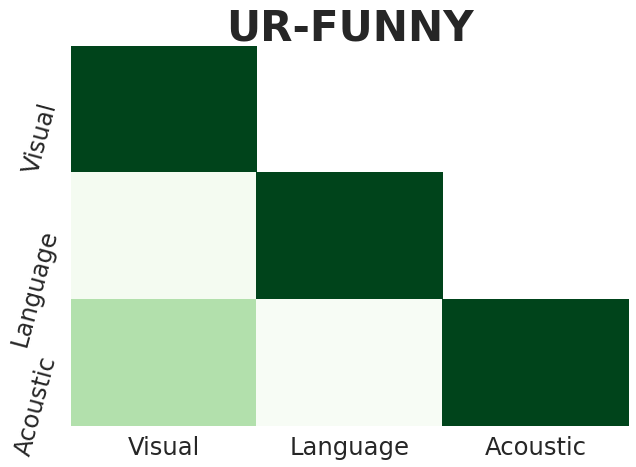}}
\subfloat{
\includegraphics[width=0.178\textwidth, height=0.16\textwidth]{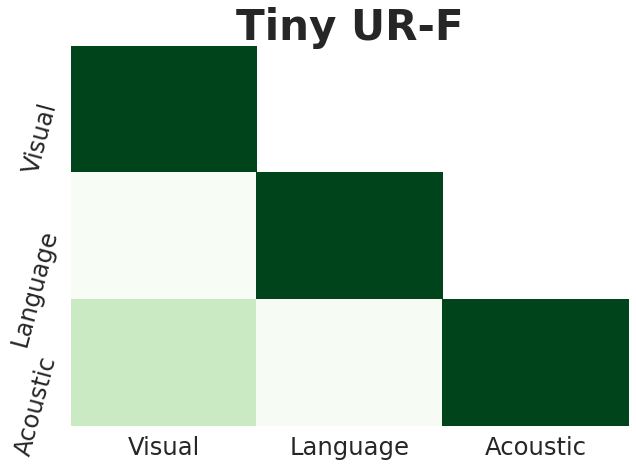}}\hspace{0.1em}

\caption{Pearson correlation calculated among the prediction outputs (dev \& test combined) of unimodal models.}
\label{fig:unimodal_pearson}
\vspace{-0.5cm}

\end{figure*}

\subsection{Ablation Study}

\noindent \textbf{Role of Modalities:} In Table \ref{tab:results_ablation}, we retrain our system after removing one modality at a time to observe the effect in performance. First, we see that the trimodal system produces the best performance compared to all bimodal/unimodal systems, indicating that each modality contains complementary information. In the case of MUStARD, we observe similar drop in performance by removing either visual or language modalities, while for UR-FUNNY, the removal of language hurts the most. Since the cameras move a lot and seldom focuses on the faces in the TED videos of UR-FUNNY, the videos in this dataset carry a lot of noise.

Next, we retrain the \model\ system using only one modality at a time to observe how much modality-specific information our unimodal encoders can capture. Following previous results, for MUStARD, both visual and language modality performs equally well, and for UR-FUNNY, the language modality proves to be the best. Audio modality alone can not yield good results because the audio features only capture the tonal-specific details of the speaker, such as pitch and intonation, which only makes sense when the transcript is present. For the MST dataset, the image and caption together perform the best. The OCR text is only present in around $57.2\%$ of the samples and thus contributes less than other modalities. Figure \ref{fig:unimodal_pearson} shows the complementarity of different modalities in each dataset. We extract the predictions from each unimodal encoder and calculate the Pearson correlation among them. The low correlation values, shown in lighter shades, indicate that each modality covers different aspects of information. The trimodal \model\ system integrates all these modalities in a unified framework and proves its efficacy.  

\noindent \textbf{Role of Intra- \& Cross-modal Attention:} Incorporating intra- and cross-modal attention significantly improves the performance, as shown in Table \ref{tab:results_ablation}. Since the incongruity, irony, and exaggeration of sarcasm and humor can generate from different modalities, cross-modal learning becomes the key, as illustrated in Figure \ref{fig:mustard_example}.


\noindent \textbf{Role of MAF:} Since the importance of three modalities can vary for different samples, we utilize an attention-based mechanism (MAF) to fuse three different sets of features. As shown in Table \ref{tab:results_ablation}, we observe that compared to a simple concatenation of the modalities, MAF improves the performance by $2-4\%$ across the five datasets. We also find that a $3$-layer MAF network is sufficient to provide the best performance.


\subsection{Interpretability of \model}

\begin{figure}[h]
\centering
\hspace*{0cm}
  \includegraphics[scale=0.09]{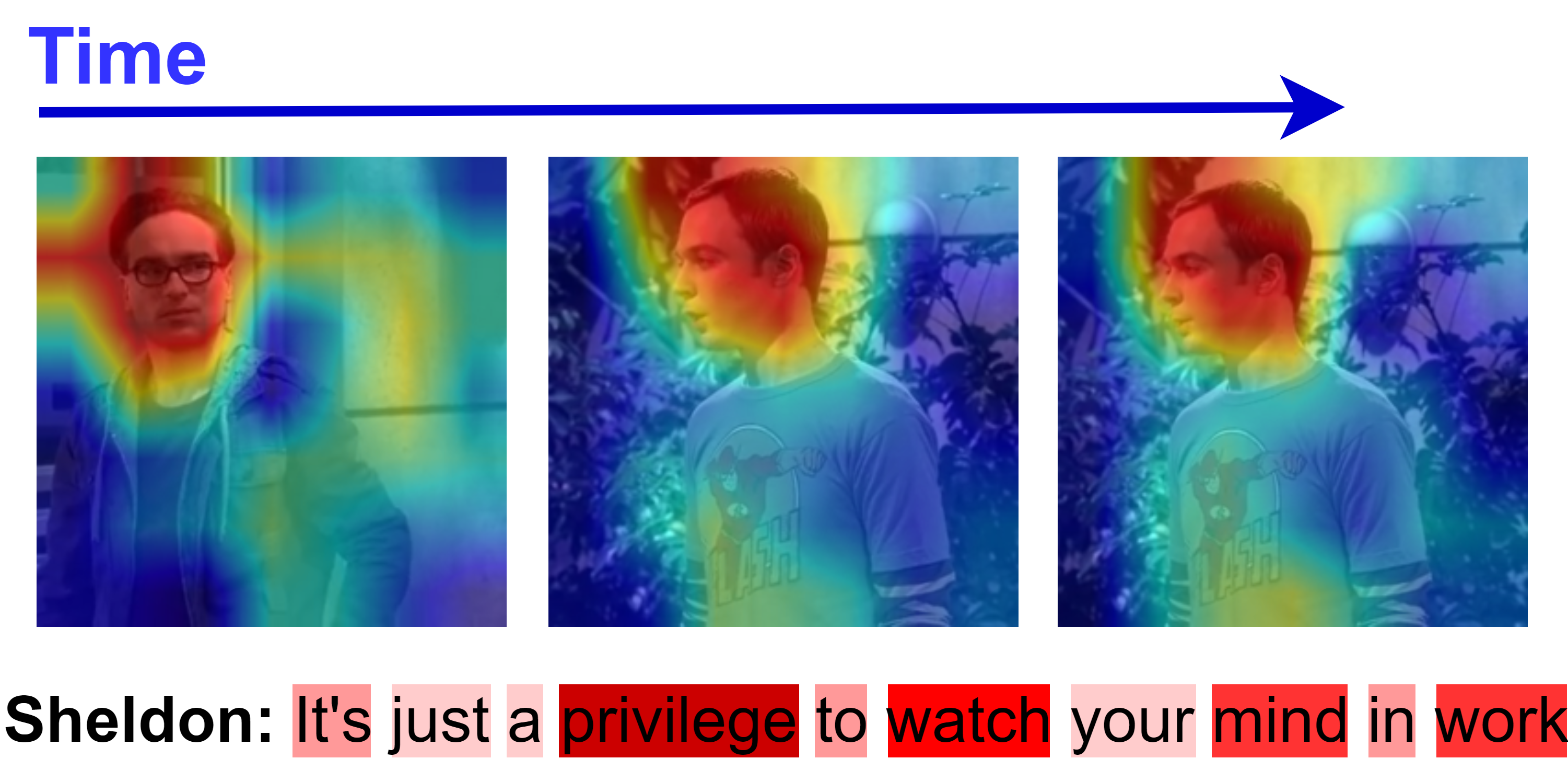}
  \caption{Visual explanations and textual attention map for a sarcastic utterance from the MUStARD dataset.}
  \label{fig:mustard_example_attention}
  \vspace{-0.5cm}
\end{figure}

\begin{figure}[h]
\centering
\hspace*{0cm}
  \includegraphics[scale=0.05]{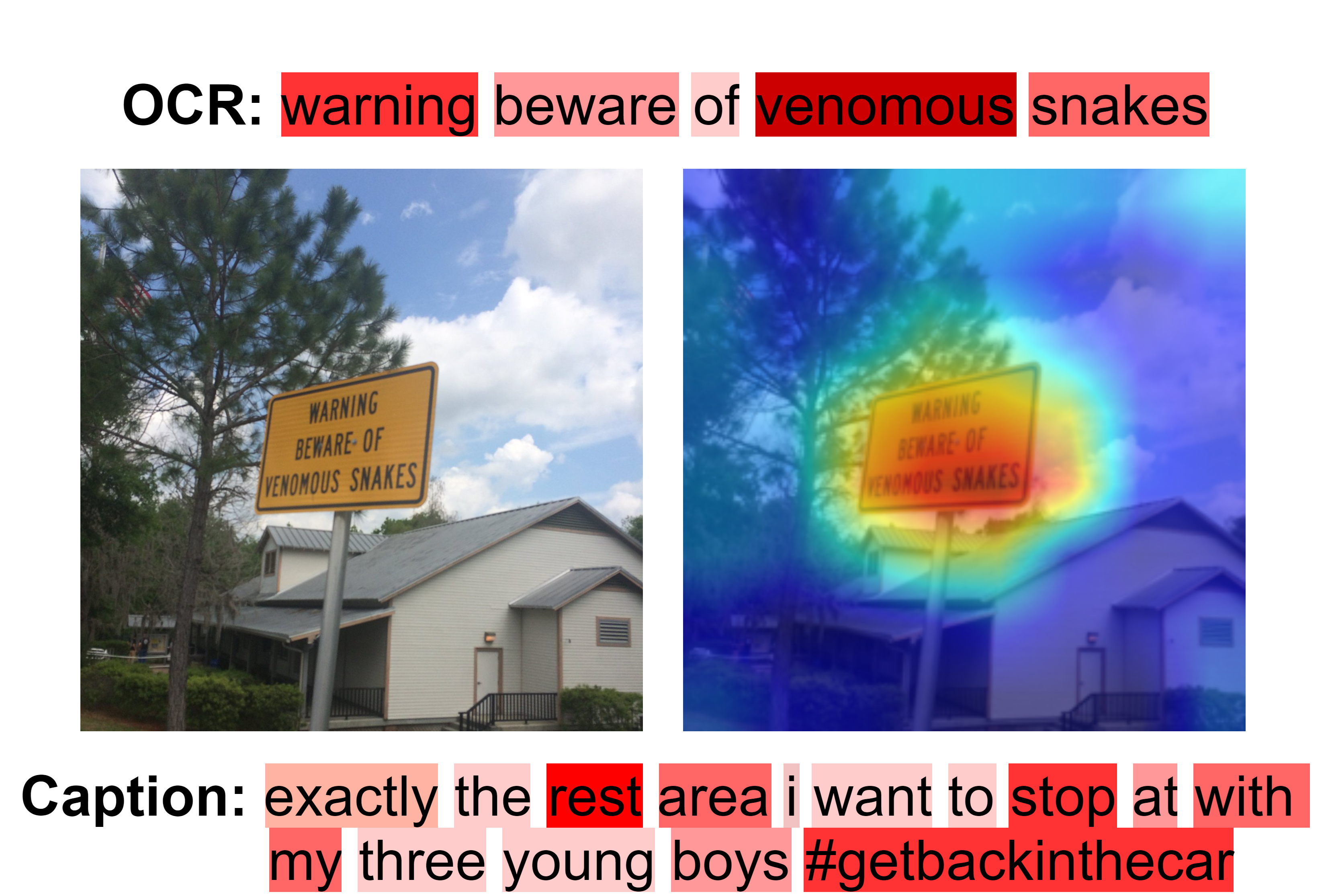}
  \caption{Visual explanations and textual attention maps for a sarcastic tweet from the MST dataset.}
  \label{fig:MST_example_attention}
  \vspace{-0.5cm}
\end{figure}

In this section, we comprehend the interpretability of \model\ by generating the visual explanations over the videos and images by using Grad-CAM \cite{selvaraju2017grad} and visualize the attention distribution over text. Figure \ref{fig:mustard_example_attention} shows three different frames of a sarcastic utterance (video id: $1\textunderscore60$) from the MUStARD dataset. \model\ focuses on the facial expression (straight face) of the speaker for all frames. In the language modality, the system focuses on words like "privilege" and thus, can detect the incongruity across the modalities. Figure \ref{fig:MST_example_attention} shows a sarcastic tweet from the MST dataset. In this sample, OCR successfully detects the text present on the image and the system perceives the irony between the OCR extracted text and the caption. Moreover, the class-discriminative regions on the image are properly identified by \model, as shown by Grad-CAM.  

\section{Conclusion}

In this paper, we deal with the task of detecting multimodal sarcasm and humor from conversational videos and image-text pairs. Capturing the intra- and inter-modal dynamics for these actions, which are highly dependent on the synchronization and aggregation of different modalities, is quite challenging with limited training data. To this end, we propose \textbf{\model}, which captures intra-modal dynamics using multi-head self-attention and cross-modal dynamics using optimal transport. Finally, multimodal attention fusion across the modalities has been performed, which further improves the performance. Experimental results on three benchmark datasets- MUStARD (video, audio, text), UR-FUNNY (video, audio, text), MST (image, text) shows \textit{2.1\%}, \textit{1.54\%} and \textit{2.34\%} performance improvements compared to the state-of-the-art using our proposed method. 

\section*{Acknowledgement}
The authors would like to thank Prof. Rama Chellappa for his helpful suggestions. Aniket Roy was supported by an ONR MURI grant N00014-20-1-2787 and Vishal M. Patel was supported by the NSF grant 1910141.

{\small
\bibliographystyle{ieee_fullname}
\bibliography{main}

\begin{thebibliography}{10}\itemsep=-1pt

\bibitem{akhtar2019multi}
Md~Shad Akhtar, Dushyant Chauhan, Deepanway Ghosal, Soujanya Poria, Asif Ekbal,
  and Pushpak Bhattacharyya.
\newblock Multi-task learning for multi-modal emotion recognition and sentiment
  analysis.
\newblock In {\em Proceedings of the 2019 Conference of the North American
  Chapter of the Association for Computational Linguistics: Human Language
  Technologies, Volume 1 (Long and Short Papers)}, pages 370--379, 2019.

\bibitem{atri2021see}
Yash~Kumar Atri, Shraman Pramanick, Vikram Goyal, and Tanmoy Chakraborty.
\newblock See, hear, read: Leveraging multimodality with guided attention for
  abstractive text summarization.
\newblock {\em Knowledge-Based Systems}, page 107152, 2021.

\bibitem{cai-etal-2019-multi}
Yitao Cai, Huiyu Cai, and Xiaojun Wan.
\newblock Multi-modal sarcasm detection in {T}witter with hierarchical fusion
  model.
\newblock In {\em Proceedings of the 57th Annual Meeting of the Association for
  Computational Linguistics}, pages 2506--2515, Florence, Italy, July 2019.
  Association for Computational Linguistics.

\bibitem{cambria2017affective}
Erik Cambria, Dipankar Das, Sivaji Bandyopadhyay, and Antonio Feraco.
\newblock Affective computing and sentiment analysis.
\newblock In {\em A practical guide to sentiment analysis}, pages 1--10.
  Springer, 2017.

\bibitem{cao2016cross}
Donglin Cao, Rongrong Ji, Dazhen Lin, and Shaozi Li.
\newblock A cross-media public sentiment analysis system for microblog.
\newblock {\em Multimedia Systems}, 22(4):479--486, 2016.

\bibitem{carreira2017quo}
Joao Carreira and Andrew Zisserman.
\newblock Quo vadis, action recognition? a new model and the kinetics dataset.
\newblock In {\em proceedings of the IEEE Conference on Computer Vision and
  Pattern Recognition}, pages 6299--6308, 2017.

\bibitem{castro-etal-2019-towards}
Santiago Castro, Devamanyu Hazarika, Ver{\'o}nica P{\'e}rez-Rosas, Roger
  Zimmermann, Rada Mihalcea, and Soujanya Poria.
\newblock Towards multimodal sarcasm detection (an {\_}{O}bviously{\_} perfect
  paper).
\newblock In {\em Proceedings of the 57th Annual Meeting of the Association for
  Computational Linguistics}, pages 4619--4629, Florence, Italy, July 2019.
  Association for Computational Linguistics.

\bibitem{chandrasekaran2016we}
Arjun Chandrasekaran, Ashwin~K Vijayakumar, Stanislaw Antol, Mohit Bansal,
  Dhruv Batra, C~Lawrence Zitnick, and Devi Parikh.
\newblock We are humor beings: Understanding and predicting visual humor.
\newblock In {\em Proceedings of the IEEE Conference on Computer Vision and
  Pattern Recognition}, pages 4603--4612, 2016.

\bibitem{chen2019complementary}
Feiyang Chen, Ziqian Luo, Yanyan Xu, and Dengfeng Ke.
\newblock Complementary fusion of multi-features and multi-modalities in
  sentiment analysis.
\newblock {\em arXiv preprint arXiv:1904.08138}, 2019.

\bibitem{chen2018humor}
Peng-Yu Chen and Von-Wun Soo.
\newblock Humor recognition using deep learning.
\newblock In {\em Proceedings of the 2018 conference of the north american
  chapter of the association for computational linguistics: Human language
  technologies, volume 2 (short papers)}, pages 113--117, 2018.

\bibitem{cortes1995support}
Corinna Cortes and Vladimir Vapnik.
\newblock Support-vector networks.
\newblock {\em Machine learning}, 20(3):273--297, 1995.

\bibitem{degottex2014covarep}
Gilles Degottex, John Kane, Thomas Drugman, Tuomo Raitio, and Stefan Scherer.
\newblock Covarep—a collaborative voice analysis repository for speech
  technologies.
\newblock In {\em 2014 ieee international conference on acoustics, speech and
  signal processing (icassp)}, pages 960--964. IEEE, 2014.

\bibitem{deng2009imagenet}
Jia Deng, Wei Dong, Richard Socher, Li-Jia Li, Kai Li, and Li Fei-Fei.
\newblock Imagenet: A large-scale hierarchical image database.
\newblock In {\em 2009 IEEE conference on computer vision and pattern
  recognition}, pages 248--255. Ieee, 2009.

\bibitem{devlin-etal-2019-bert}
Jacob Devlin, Ming-Wei Chang, Kenton Lee, and Kristina Toutanova.
\newblock {BERT}: Pre-training of deep bidirectional transformers for language
  understanding.
\newblock In {\em Proceedings of the 2019 Conference of the North {A}merican
  Chapter of the Association for Computational Linguistics: Human Language
  Technologies, Volume 1 (Long and Short Papers)}, pages 4171--4186,
  Minneapolis, Minnesota, June 2019. Association for Computational Linguistics.

\bibitem{drugman2011joint}
Thomas Drugman and Abeer Alwan.
\newblock Joint robust voicing detection and pitch estimation based on residual
  harmonics.
\newblock In {\em INTERSPEECH 2011, Twelfth Annual Conference of the
  International Speech Communication Association}, 2011.

\bibitem{drugman2011detection}
Thomas Drugman, Mark Thomas, Jon Gudnason, Patrick Naylor, and Thierry Dutoit.
\newblock Detection of glottal closure instants from speech signals: A
  quantitative review.
\newblock {\em IEEE Transactions on Audio, Speech, and Language Processing},
  20(3):994--1006, 2011.

\bibitem{flamary2021pot}
R{\'e}mi Flamary, Nicolas Courty, Alexandre Gramfort, Mokhtar~Z. Alaya,
  Aur{\'e}lie Boisbunon, Stanislas Chambon, Laetitia Chapel, Adrien Corenflos,
  Kilian Fatras, Nemo Fournier, L{\'e}o Gautheron, Nathalie~T.H. Gayraud,
  Hicham Janati, Alain Rakotomamonjy, Ievgen Redko, Antoine Rolet, Antony
  Schutz, Vivien Seguy, Danica~J. Sutherland, Romain Tavenard, Alexander Tong,
  and Titouan Vayer.
\newblock Pot: Python optimal transport.
\newblock {\em Journal of Machine Learning Research}, 22(78):1--8, 2021.

\bibitem{gu2018hybrid}
Yue Gu, Kangning Yang, Shiyu Fu, Shuhong Chen, Xinyu Li, and Ivan Marsic.
\newblock Hybrid attention based multimodal network for spoken language
  classification.
\newblock In {\em ACL}, volume 2018, page 2379, 2018.

\bibitem{han2021bi}
Wei Han, Hui Chen, Alexander Gelbukh, Amir Zadeh, Louis-philippe Morency, and
  Soujanya Poria.
\newblock Bi-bimodal modality fusion for correlation-controlled multimodal
  sentiment analysis.
\newblock {\em 23rd ACM International Conference on Multimodal Interaction
  (ICMI)}, 2021.

\bibitem{hasan-etal-2019-ur}
Md~Kamrul Hasan, Wasifur Rahman, AmirAli Bagher~Zadeh, Jianyuan Zhong,
  Md~Iftekhar Tanveer, Louis-Philippe Morency, and Mohammed~(Ehsan) Hoque.
\newblock {UR}-{FUNNY}: A multimodal language dataset for understanding humor.
\newblock In {\em Proceedings of the 2019 Conference on Empirical Methods in
  Natural Language Processing and the 9th International Joint Conference on
  Natural Language Processing (EMNLP-IJCNLP)}, pages 2046--2056, Hong Kong,
  China, Nov. 2019. Association for Computational Linguistics.

\bibitem{hazarika2020misa}
Devamanyu Hazarika, Roger Zimmermann, and Soujanya Poria.
\newblock Misa: Modality-invariant and-specific representations for multimodal
  sentiment analysis.
\newblock In {\em Proceedings of the 28th ACM International Conference on
  Multimedia}, pages 1122--1131, 2020.

\bibitem{he2016deep}
Kaiming He, Xiangyu Zhang, Shaoqing Ren, and Jian Sun.
\newblock Deep residual learning for image recognition.
\newblock In {\em Proceedings of the IEEE conference on computer vision and
  pattern recognition}, pages 770--778, 2016.

\bibitem{howard2018universal}
Jeremy Howard and Sebastian Ruder.
\newblock Universal language model fine-tuning for text classification.
\newblock In {\em Proceedings of the 56th Annual Meeting of the Association for
  Computational Linguistics (Volume 1: Long Papers)}, pages 328--339, 2018.

\bibitem{huang2020multimodal}
Jian Huang, Jianhua Tao, Bin Liu, Zheng Lian, and Mingyue Niu.
\newblock Multimodal transformer fusion for continuous emotion recognition.
\newblock In {\em ICASSP 2020-2020 IEEE International Conference on Acoustics,
  Speech and Signal Processing (ICASSP)}, pages 3507--3511. IEEE, 2020.

\bibitem{kant2020spatially}
Yash Kant, Dhruv Batra, Peter Anderson, Alexander Schwing, Devi Parikh, Jiasen
  Lu, and Harsh Agrawal.
\newblock Spatially aware multimodal transformers for textvqa.
\newblock In {\em 16th European Conference on Computer Vision, ECCV 2020},
  pages 715--732. Springer Science and Business Media Deutschland GmbH, 2020.

\bibitem{kay2017kinetics}
Will Kay, Joao Carreira, Karen Simonyan, Brian Zhang, Chloe Hillier, Sudheendra
  Vijayanarasimhan, Fabio Viola, Tim Green, Trevor Back, Paul Natsev, et~al.
\newblock The kinetics human action video dataset.
\newblock {\em CoRR}, 2017.

\bibitem{kiela2020supervised}
Douwe Kiela, Suvrat Bhooshan, Hamed Firooz, Ethan Perez, and Davide Testuggine.
\newblock Supervised multimodal bitransformers for classifying images and text.
\newblock {\em arxiv:1909.02950}, 2020.

\bibitem{kolchinski2018representing}
Y~Alex Kolchinski and Christopher Potts.
\newblock Representing social media users for sarcasm detection.
\newblock In {\em Proceedings of the 2018 Conference on Empirical Methods in
  Natural Language Processing}, pages 1115--1121, 2018.

\bibitem{lefcourt2012humor}
Herbert~M Lefcourt and Rod~A Martin.
\newblock {\em Humor and life stress: Antidote to adversity}.
\newblock Springer Science \& Business Media, 2012.

\bibitem{liu2018modeling}
Lizhen Liu, Donghai Zhang, and Wei Song.
\newblock Modeling sentiment association in discourse for humor recognition.
\newblock In {\em Proceedings of the 56th Annual Meeting of the Association for
  Computational Linguistics (Volume 2: Short Papers)}, pages 586--591, 2018.

\bibitem{liu2019roberta}
Yinhan Liu, Myle Ott, Naman Goyal, Jingfei Du, Mandar Joshi, Danqi Chen, Omer
  Levy, Mike Lewis, Luke Zettlemoyer, and Veselin Stoyanov.
\newblock Roberta: A robustly optimized bert pretraining approach.
\newblock {\em arXiv preprint arXiv:1907.11692}, 2019.

\bibitem{loshchilov2018decoupled}
Ilya Loshchilov and Frank Hutter.
\newblock Decoupled weight decay regularization.
\newblock In {\em International Conference on Learning Representations (ICLR)},
  2018.

\bibitem{lu2019vilbert}
Jiasen Lu, Dhruv Batra, Devi Parikh, and Stefan Lee.
\newblock Vilbert: pretraining task-agnostic visiolinguistic representations
  for vision-and-language tasks.
\newblock In {\em Proceedings of the 33rd International Conference on Neural
  Information Processing Systems}, pages 13--23, 2019.

\bibitem{mialon2021trainable}
Gr{\'e}goire Mialon, Dexiong Chen, Alexandre d'Aspremont, and Julien Mairal.
\newblock A trainable optimal transport embedding for feature aggregation and
  its relationship to attention.
\newblock In {\em ICLR 2021-The Ninth International Conference on Learning
  Representations}, 2021.

\bibitem{mikolov2013efficient}
Tomas Mikolov, Kai Chen, Greg Corrado, and Jeffrey Dean.
\newblock Efficient estimation of word representations in vector space.
\newblock {\em arXiv preprint arXiv:1301.3781}, 2013.

\bibitem{olson1977utterance}
David Olson.
\newblock From utterance to text: The bias of language in speech and writing.
\newblock {\em Harvard educational review}, 47(3):257--281, 1977.

\bibitem{pan2020modeling}
Hongliang Pan, Zheng Lin, Peng Fu, Yatao Qi, and Weiping Wang.
\newblock Modeling intra and inter-modality incongruity for multi-modal sarcasm
  detection.
\newblock In {\em Proceedings of the 2020 Conference on Empirical Methods in
  Natural Language Processing: Findings}, pages 1383--1392, 2020.

\bibitem{paszke2019pytorch}
Adam Paszke, Sam Gross, Francisco Massa, Adam Lerer, James Bradbury, Gregory
  Chanan, Trevor Killeen, Zeming Lin, Natalia Gimelshein, Luca Antiga, et~al.
\newblock Pytorch: An imperative style, high-performance deep learning library.
\newblock {\em Advances in neural information processing systems},
  32:8026--8037, 2019.

\bibitem{pennington2014glove}
Jeffrey Pennington, Richard Socher, and Christopher~D Manning.
\newblock Glove: Global vectors for word representation.
\newblock In {\em Proceedings of the 2014 conference on empirical methods in
  natural language processing (EMNLP)}, pages 1532--1543, 2014.

\bibitem{peters2018deep}
Matthew Peters, Mark Neumann, Mohit Iyyer, Matt Gardner, Christopher Clark,
  Kenton Lee, and Luke Zettlemoyer.
\newblock Deep contextualized word representations.
\newblock In {\em Proceedings of the 2018 Conference of the North American
  Chapter of the Association for Computational Linguistics: Human Language
  Technologies, Volume 1 (Long Papers)}, pages 2227--2237, 2018.

\bibitem{pham2019found}
Hai Pham, Paul~Pu Liang, Thomas Manzini, Louis-Philippe Morency, and
  Barnab{\'a}s P{\'o}czos.
\newblock Found in translation: Learning robust joint representations by cyclic
  translations between modalities.
\newblock In {\em Proceedings of the AAAI Conference on Artificial
  Intelligence}, volume~33, pages 6892--6899, 2019.

\bibitem{poria2017review}
Soujanya Poria, Erik Cambria, Rajiv Bajpai, and Amir Hussain.
\newblock A review of affective computing: From unimodal analysis to multimodal
  fusion.
\newblock {\em Information Fusion}, 37:98--125, 2017.

\bibitem{poria2016convolutional}
Soujanya Poria, Iti Chaturvedi, Erik Cambria, and Amir Hussain.
\newblock Convolutional mkl based multimodal emotion recognition and sentiment
  analysis.
\newblock In {\em 2016 IEEE 16th international conference on data mining
  (ICDM)}, pages 439--448. IEEE, 2016.

\bibitem{pramanick2021exercise}
Shraman Pramanick, Md~Shad Akhtar, and Tanmoy Chakraborty.
\newblock Exercise? {I} thought you said 'extra fries’: Leveraging sentence
  demarcations and multi-hop attention for meme affect analysis.
\newblock In {\em Proceedings of the International AAAI Conference on Web and
  Social Media}, volume~15, pages 513--524, 2021.

\bibitem{pramanick2021detecting}
Shraman Pramanick, Dimitar Dimitrov, Rituparna Mukherjee, Shivam Sharma,
  Md~Shad Akhtar, Preslav Nakov, and Tanmoy Chakraborty.
\newblock Detecting harmful memes and their targets.
\newblock In {\em Findings of the Association for Computational Linguistics:
  ACL-IJCNLP 2021}, pages 2783--2796, 2021.

\bibitem{pramanick2021momenta}
Shraman Pramanick, Shivam Sharma, Dimitar Dimitrov, Md~Shad Akhtar, Preslav
  Nakov, and Tanmoy Chakraborty.
\newblock Momenta: A multimodal framework for detecting harmful memes and their
  targets.
\newblock {\em arXiv preprint arXiv:2109.05184}, 2021.

\bibitem{rahman2020integrating}
Wasifur Rahman, Md~Kamrul Hasan, Sangwu Lee, Amir Zadeh, Chengfeng Mao,
  Louis-Philippe Morency, and Ehsan Hoque.
\newblock Integrating multimodal information in large pretrained transformers.
\newblock In {\em Proceedings of the conference. Association for Computational
  Linguistics. Meeting}, volume 2020, page 2359. NIH Public Access, 2020.

\bibitem{riloff2013sarcasm}
Ellen Riloff, Ashequl Qadir, Prafulla Surve, Lalindra De~Silva, Nathan Gilbert,
  and Ruihong Huang.
\newblock Sarcasm as contrast between a positive sentiment and negative
  situation.
\newblock In {\em Proceedings of the 2013 conference on empirical methods in
  natural language processing}, pages 704--714, 2013.

\bibitem{selvaraju2017grad}
Ramprasaath~R Selvaraju, Michael Cogswell, Abhishek Das, Ramakrishna Vedantam,
  Devi Parikh, and Dhruv Batra.
\newblock Grad-cam: Visual explanations from deep networks via gradient-based
  localization.
\newblock In {\em Proceedings of the IEEE international conference on computer
  vision}, pages 618--626, 2017.

\bibitem{sharma2018conceptual}
Piyush Sharma, Nan Ding, Sebastian Goodman, and Radu Soricut.
\newblock Conceptual captions: A cleaned, hypernymed, image alt-text dataset
  for automatic image captioning.
\newblock In {\em Proceedings of the 56th Annual Meeting of the Association for
  Computational Linguistics (Volume 1: Long Papers)}, pages 2556--2565, 2018.

\bibitem{sun2019videobert}
Chen Sun, Austin Myers, Carl Vondrick, Kevin Murphy, and Cordelia Schmid.
\newblock Videobert: A joint model for video and language representation
  learning.
\newblock In {\em Proceedings of the IEEE/CVF International Conference on
  Computer Vision}, pages 7464--7473, 2019.

\bibitem{sun2020learning}
Zhongkai Sun, Prathusha Sarma, William Sethares, and Yingyu Liang.
\newblock Learning relationships between text, audio, and video via deep
  canonical correlation for multimodal language analysis.
\newblock In {\em Proceedings of the AAAI Conference on Artificial
  Intelligence}, volume~34, pages 8992--8999, 2020.

\bibitem{tepperman2006yeah}
Joseph Tepperman, David Traum, and Shrikanth Narayanan.
\newblock " yeah right": Sarcasm recognition for spoken dialogue systems.
\newblock In {\em Ninth international conference on spoken language
  processing}, 2006.

\bibitem{tsai-etal-2019-multimodal}
Yao-Hung~Hubert Tsai, Shaojie Bai, Paul~Pu Liang, J.~Zico Kolter,
  Louis-Philippe Morency, and Ruslan Salakhutdinov.
\newblock Multimodal transformer for unaligned multimodal language sequences.
\newblock In {\em Proceedings of the 57th Annual Meeting of the Association for
  Computational Linguistics}, pages 6558--6569, Florence, Italy, July 2019.
  Association for Computational Linguistics.

\bibitem{vartabedian1993humor}
Robert~A Vartabedian and Laurel~Klinger Vartabedian.
\newblock Humor in the workplace: A communication challenge.
\newblock 1993.

\bibitem{vaswani2017attention}
Ashish Vaswani, Noam Shazeer, Niki Parmar, Jakob Uszkoreit, Llion Jones,
  Aidan~N Gomez, {\L}ukasz Kaiser, and Illia Polosukhin.
\newblock Attention is all you need.
\newblock In {\em Advances in neural information processing systems}, pages
  5998--6008, 2017.

\bibitem{villani2008optimal}
C Villani.
\newblock Optimal transport, old and new. notes for the 2005 saint-flour summer
  school.
\newblock {\em Grundlehren der mathematischen Wissenschaften [Fundamental
  Principles of Mathematical Sciences]. Springer}, 2008.

\bibitem{wang2019words}
Yansen Wang, Ying Shen, Zhun Liu, Paul~Pu Liang, Amir Zadeh, and Louis-Philippe
  Morency.
\newblock Words can shift: Dynamically adjusting word representations using
  nonverbal behaviors.
\newblock In {\em Proceedings of the AAAI Conference on Artificial
  Intelligence}, volume~33, pages 7216--7223, 2019.

\bibitem{wei2020multi}
Xi Wei, Tianzhu Zhang, Yan Li, Yongdong Zhang, and Feng Wu.
\newblock Multi-modality cross attention network for image and sentence
  matching.
\newblock In {\em Proceedings of the IEEE/CVF conference on computer vision and
  pattern recognition}, pages 10941--10950, 2020.

\bibitem{woodland2011context}
Jennifer Woodland and Daniel Voyer.
\newblock Context and intonation in the perception of sarcasm.
\newblock {\em Metaphor and Symbol}, 26(3):227--239, 2011.

\bibitem{xu2020reasoning}
Nan Xu, Zhixiong Zeng, and Wenji Mao.
\newblock Reasoning with multimodal sarcastic tweets via modeling
  cross-modality contrast and semantic association.
\newblock In {\em Proceedings of the 58th Annual Meeting of the Association for
  Computational Linguistics}, pages 3777--3786, 2020.

\bibitem{yang2015humor}
Diyi Yang, Alon Lavie, Chris Dyer, and Eduard Hovy.
\newblock Humor recognition and humor anchor extraction.
\newblock In {\em Proceedings of the 2015 Conference on Empirical Methods in
  Natural Language Processing}, pages 2367--2376, 2015.

\bibitem{yang2019xlnet}
Zhilin Yang, Zihang Dai, Yiming Yang, Jaime Carbonell, Russ~R Salakhutdinov,
  and Quoc~V Le.
\newblock Xlnet: Generalized autoregressive pretraining for language
  understanding.
\newblock {\em Advances in neural information processing systems}, 32, 2019.

\bibitem{zadeh2017tensor}
Amir Zadeh, Minghai Chen, Soujanya Poria, Erik Cambria, and Louis-Philippe
  Morency.
\newblock Tensor fusion network for multimodal sentiment analysis.
\newblock In {\em Proceedings of the 2017 Conference on Empirical Methods in
  Natural Language Processing}, pages 1103--1114, 2017.

\bibitem{zadeh2018memory}
Amir Zadeh, Paul~Pu Liang, Navonil Mazumder, Soujanya Poria, Erik Cambria, and
  Louis-Philippe Morency.
\newblock Memory fusion network for multi-view sequential learning.
\newblock In {\em Proceedings of the AAAI Conference on Artificial
  Intelligence}, volume~32, 2018.

\bibitem{zadeh2018multimodal}
AmirAli~Bagher Zadeh, Paul~Pu Liang, Soujanya Poria, Erik Cambria, and
  Louis-Philippe Morency.
\newblock Multimodal language analysis in the wild: Cmu-mosei dataset and
  interpretable dynamic fusion graph.
\newblock In {\em Proceedings of the 56th Annual Meeting of the Association for
  Computational Linguistics (Volume 1: Long Papers)}, pages 2236--2246, 2018.

\end{thebibliography}
}

\newpage
\thispagestyle{plain}
\makeatletter
\twocolumn[\LARGE \bf \centering Multimodal Learning using Optimal Transport \\ for Sarcasm and Humor Detection \\ (Supplementary) \par \bigskip
 ]
\appendix

\counterwithin{figure}{section}
\numberwithin{table}{section}

In this supplementary document, we provide additional details on the hyperparameter values, baselines and illustrate additional qualitative results.

\section{Implementation Details \& Hyperparameter Values}

In Table \ref{tab:hyperparameters}, we furnish the details of hyperparameters used during training. Grid search is performed on batch size, learning rate, and the number of attention-heads to find the best hyperparameter configuration. The model is evaluated after each $10$ steps on the dev set and the best model was taken to be evaluated on the test set. We use the weighted Adam optimizer \cite{loshchilov2018decoupled} with step learning rate scheduler and fixed $2000$ warmup steps for optimization without gradient clipping.

\begin{table}[h]
\centering 

\resizebox{\columnwidth}{!}
{

\begin{tabular}{p{4.3cm}|l|l}

\textbf{Hyper-parameter} & \textbf{Notation} & \textbf{Value} \\
\hline

\hline
{\#dim} for Dense layers in MAF & - & $[64, 8, 4, 1]$ \\
{\#dim} for Classification FC & - & $[192, 32, 8, 2]$ \\
\hline
\multicolumn{3}{c}{Intra-modality Attention} \\ \hline
{\#heads} & $k$ & 80 \\
{\#dim} in Equation \ref{equ:weight_self_attention} & r & 30 \\ \hline

\multicolumn{3}{c}{Cross-modality Attention} \\ \hline
Shared feature sequence length & $L_{uni}$ & 100 \\
Shared feature dimension & $d_{uni}$ & 64 \\ \hline

\multicolumn{3}{c}{Training} \\ \hline
{Batch-size} & - & $64$ \\
{Epochs} & $N$ & $300$ \\ 
{Optimizer} & - & AdamW \\ 
{Loss} & - & BCE \\ 
{Learning-rate} & $\alpha$ & $0.005$ \\
{Learning-rate-decay (/10K$ iter$)} & - & $1$e-$4$ \\
Betas & [$\beta_1$, $\beta_2$] & [$0.9$, $0.999$] \\
Weight Decay & - & 0.01 \\
\hline
\multicolumn{3}{c}{Class weights for imbalanced training data} \\ \hline
\multicolumn{2}{l|}{Training on MST $[w_{sar}, w_{non-sar}]$} & $[1.2, 1]$ \\
 \hline
 
\hline
\end{tabular}}
\caption{Hyper-parameters of \model.}
\label{tab:hyperparameters}
\end{table}

\begin{figure*}[t]
\centering
\hspace*{0cm}
  \includegraphics[scale=0.135]{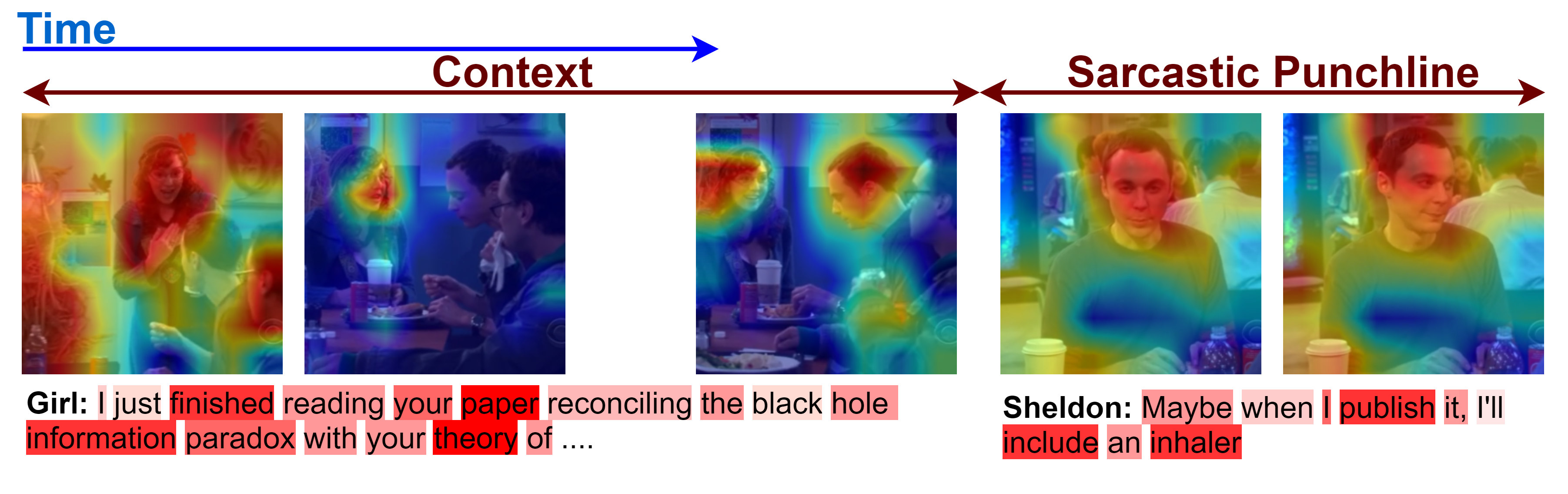}
  \caption{Visual explanations and textual attention maps of the context and punchline of a sarcastic utterance from the MUStARD dataset.}
  \label{fig:mustard_example_additional}
\end{figure*}

\begin{figure*}[t]
\centering
\hspace*{0cm}
  \includegraphics[scale=0.135]{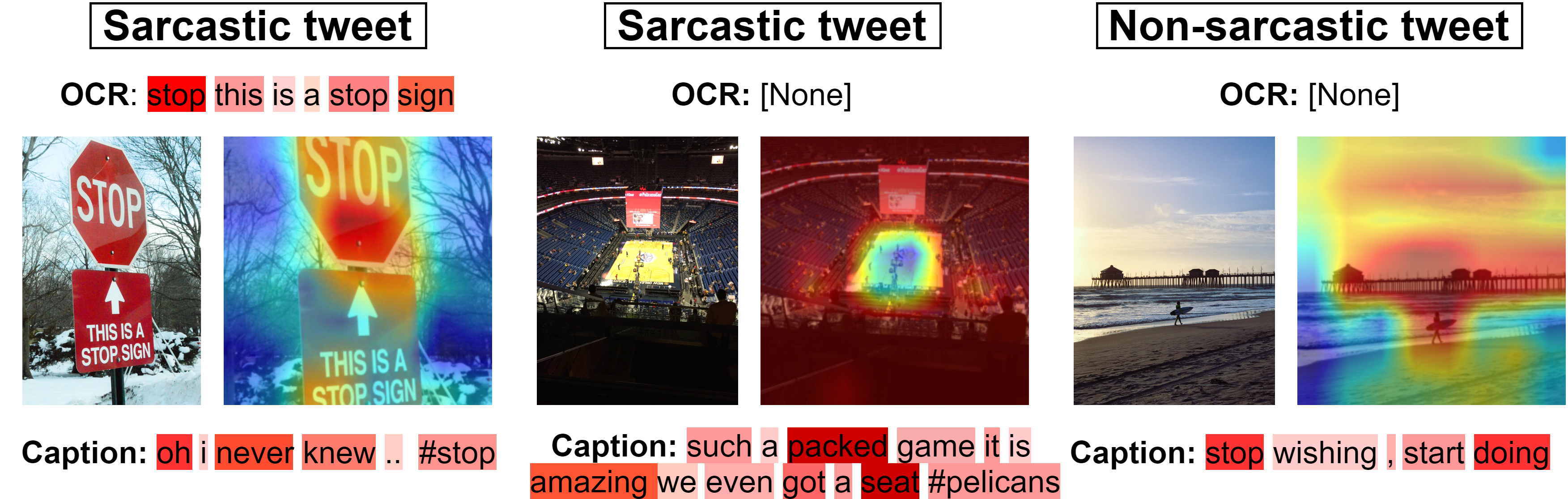}
  \caption{Visual explanations and textual attention maps of two sarcastic tweets and one non-sarcastic tweet from the MST dataset.}
  \label{fig:mst_example_additional}
\end{figure*}

\section{Baselines}
Numerous methods have been proposed in the literature for multimodal sentiment analysis, but none of them has generalized their systems both on videos and images. Therefore, we chose different baselines for videos and images. Furthermore, we remove various modalities and modules at a time from our proposed \model\ system to observe the effect in performance. Baselines on MUStARD and UR-FUNNY include:

\begin{itemize}[leftmargin= 0.1in]
\item \textbf{Support Vector Machines (SVM)} \cite{cortes1995support} was used as the baseline model for MUStARD dataset \cite{castro-etal-2019-towards}. However, they extracted individual frames from the videos and used ResNet \cite{he2016deep} for extracting framewise visual features which does not take account the temporal dynamics of a video. We used I3D \cite{carreira2017quo} for visual features, and retrain an SVM model with same configuration.
\item \textbf{DFF-ATMF} \cite{chen2019complementary} is a simple multimodal system which concatenates the unimodal features using an attention based modality fusion.
\item \textbf{CIM-MTL} \cite{akhtar2019multi} leverage the interdependence of two related tasks (sentiment and emotion) in improving each others performance using an effective multi-modal attention framework.
\item \textbf{Tensor Fusion Network (TFN)} \cite{zadeh2017tensor} models intra-modality and inter-modality dynamics concurrently with local feature extraction network and $3$-fold Cartesian product.
\item \textbf{Contextual Memory Fusion Network (CMFN)} \cite{hasan-etal-2019-ur} proposes uni- and multimodal context networks that consider preceding utterances and performs fusion using the MFN model as its backbone. Originally, MFN \cite{zadeh2018memory} is a multi-view gated memory network that stores intra- and cross-modal utterance interactions in its memories. 
\item \textbf{MISA} \cite{hazarika2020misa} is flexible multimodal learning framework that emphasizes on multimodal representation learning as a pre-cursor to multimodal fusion. 
\item \textbf{Bi-Bimodal Fusion Network (BBFN)} \cite{han2021bi} is an end-to-end network that performs fusion (relevance increment) and separation (difference increment) on pairwise modality representations, where the two parts are trained simultaneously such that the competition between them is simulated.
\item \textbf{MAG-Transformer} \cite{rahman2020integrating} introduced Multimodal Adaption Gate (MAG) to fuse acoustic and visual information in pre-trained language transformers. Due to its superior performance on various multimodal sentiment analysis datasets, we ran MAG-XLNet on MUStARD and UR-FUNNY and it yields best baseline result on both datasets.
\end{itemize}
Baselines on MST include:
\begin{itemize}[leftmargin= 0.1in]
\item \textbf{Concat BERT} concatenates the features extracted by pre-trained unimodal ResNet-152 \cite{he2016deep} and Text BERT \cite{devlin-etal-2019-bert} and uses a simple perceptron as the classifier.
\item \textbf{}\textbf{Supervised Multimodal Bitransformer (MMBT)} \cite{kiela2020supervised} captures the intra-modal and inter-modal dynamics within various input modalities.
\item \textbf{Vision and Language BERT (ViLBERT)} \cite{lu2019vilbert}, trained on an intermediate multimodal objective (Conceptual Captions) \cite{sharma2018conceptual}, is a strong model with task-agnostic joint representation of image + text.
\item \textbf{Hierarchical Fusion Model (HFM)} \cite{cai-etal-2019-multi} is similar to Concat BERT, but uses an hierarchical attention fusion mechanism for obtaining the multimodal feature representation.
\item \textbf{D\&R Net} \cite{xu2020reasoning} models cross-modality contrast and semantic association using Decomposition and Relation, where the decomposition network represents the commonality and discrepancy between image and text, and the relation network models the semantic association in cross-modality context. 
\item \textbf{MsdBERT} \cite{pan2020modeling} is the SOTA on MST dataset. This system utilizes a co-attention network to exploit intra- and inter-modality incongruity between text and image.  

\section{Additional Qualitative Results}

In this section, we illustrate a few more visual explanations over videos and images by using Grad-CAM \cite{selvaraju2017grad} and plot the corresponding textual attention maps as generated by the multi-head attention layer of \model. Figure \ref{fig:mustard_example_additional} shows five different frames from the context and punchline of a sarcastic utterance (video id: $1 \textunderscore 1466$) from MUStARD dataset. \model\ focuses on the facial expression of the speaker in every frame. In the language modality, incongruous words to the speaker's expression are attended, and thus the system detects sarcasm. Figure \ref{fig:mst_example_additional} shows two sarcastic tweets and one non-sarcastic tweet from MST dataset. In the first tweet, the model attends the words "this is a stop sign" from OCR text and "I never knew" from the caption, and recognizes the writer's joking tone. Again, in the second example, \model\ attends to the regions in image indicating lots of empty seats, which forms contradiction with the text ”packed game". In the non-sarcastic tweet, no such incongruity is present.  

\end{itemize}

\end{document}